\documentclass{article}

\usepackage{PRIMEarxiv}

\usepackage[utf8]{inputenc} % allow utf-8 input
\usepackage[T1]{fontenc}    % use 8-bit T1 fonts
\usepackage{hyperref}       % hyperlinks
\usepackage{url}            % simple URL typesetting
\usepackage{booktabs}       % professional-quality tables
\usepackage{amsfonts}       % blackboard math symbols
\usepackage{nicefrac}       % compact symbols for 1/2, etc.
\usepackage{microtype}      % microtypography
\usepackage{lipsum}
\usepackage{fancyhdr}       % header
\usepackage{graphicx}       % graphics
\graphicspath{{media/}}     % organize your images and other figures under media/ folder

\usepackage{algorithm}
\usepackage{algorithmic}

\usepackage{times}
\usepackage{latexsym}
\usepackage{xcolor}
\usepackage{soul}
\usepackage{multirow}
\usepackage{booktabs}
\usepackage{subcaption}
\usepackage[T1]{fontenc}
\usepackage[utf8]{inputenc}
\usepackage{microtype}
\usepackage{inconsolata}

\newcommand{\NTA}{{}\texttt{\color{blue} NTA}}
\newcommand{\YTA}{{}\texttt{\color{red} YTA}}
\newcommand{\AITA}{\emph{r/AmITheAsshole}}

\newcommand{\hlc}[2][yellow]{{%
    \colorlet{foo}{#1}%
    \sethlcolor{foo}\hl{#2}}%
}

\title{Author as Character and Narrator: Deconstructing Personal Narratives from the {\em r/AmITheAsshole} Reddit Community
}

\author{
  Salvatore Giorgi, Ke Zhao, Alexander H. Feng, Lara J. Martin \\
  Department of Computer and Information Science \\
  University of Pennsylvania \\
  Philadelphia, PA\\
  \texttt{sgiorgi@sas.upenn.edu}, cocozhao321@gmail.com, \{ahfeng, laramar\}@seas.upenn.edu  \\
}

\begin{document}
\maketitle

\begin{abstract}
In the \AITA~subreddit, people anonymously share first person narratives that contain some moral dilemma or conflict and ask the community to judge who is at fault (i.e., who is ``the asshole''). 
In general, first person narratives are a unique storytelling domain where the author is the narrator (the person telling the story) but can also be a character (the person living the story) and, thus, the author has two distinct voices presented in the story. 
In this study, we identify linguistic and narrative features associated with the author as the character or as a narrator.
We use these features to answer the following questions: (1) what makes an asshole character and (2) what makes an asshole narrator?
We extract both Author-as-Character features (e.g., demographics, narrative event chain, and emotional arc) and Author-as-Narrator features (i.e., the style and emotion of the story as a whole) in order to identify which aspects of the narrative are correlated with the final moral judgment. 
Our work shows that ``assholes'' as Characters frame themselves as lacking agency with a more positive personal arc, while ``assholes'' as Narrators will tell emotional and opinionated stories.
\end{abstract}

\section{Introduction}

When you read a story, you might identify with the characters and their dilemmas and not realize the biases behind the person telling the story. Readers might fail to consider the author as the narrator until external events bring the author's opinions to light, giving them extra information as they reread and reinterpret the story.
On one end, there is a clear difference between character and narrator. Consider the world of fan fiction, where fans of a particular work will tell their own stories using the same characters as the original work. This divide is exacerbated when the original creator's and the fan fiction writers' values diverge (e.g., J.K. Rowling's {\em Harry Potter} vs {\em Harry Potter} fan fiction by LGBTQ+ authors; Duggan, 2022~\cite{Duggan2022}).

On the other end of the spectrum, the characters and the narrator are intertwined. This is especially relevant when the author is both a character in the story and the narrator, making it difficult to differentiate between what moral values the author has vs what moral values the character has.
These types of stories are found in autobiographies and memoirs, but they can also be found on the internet in the form of subreddits such as \AITA.

\begin{figure}[!tb]
\centering
\minipage{.5\columnwidth}
  \includegraphics[width=\linewidth]{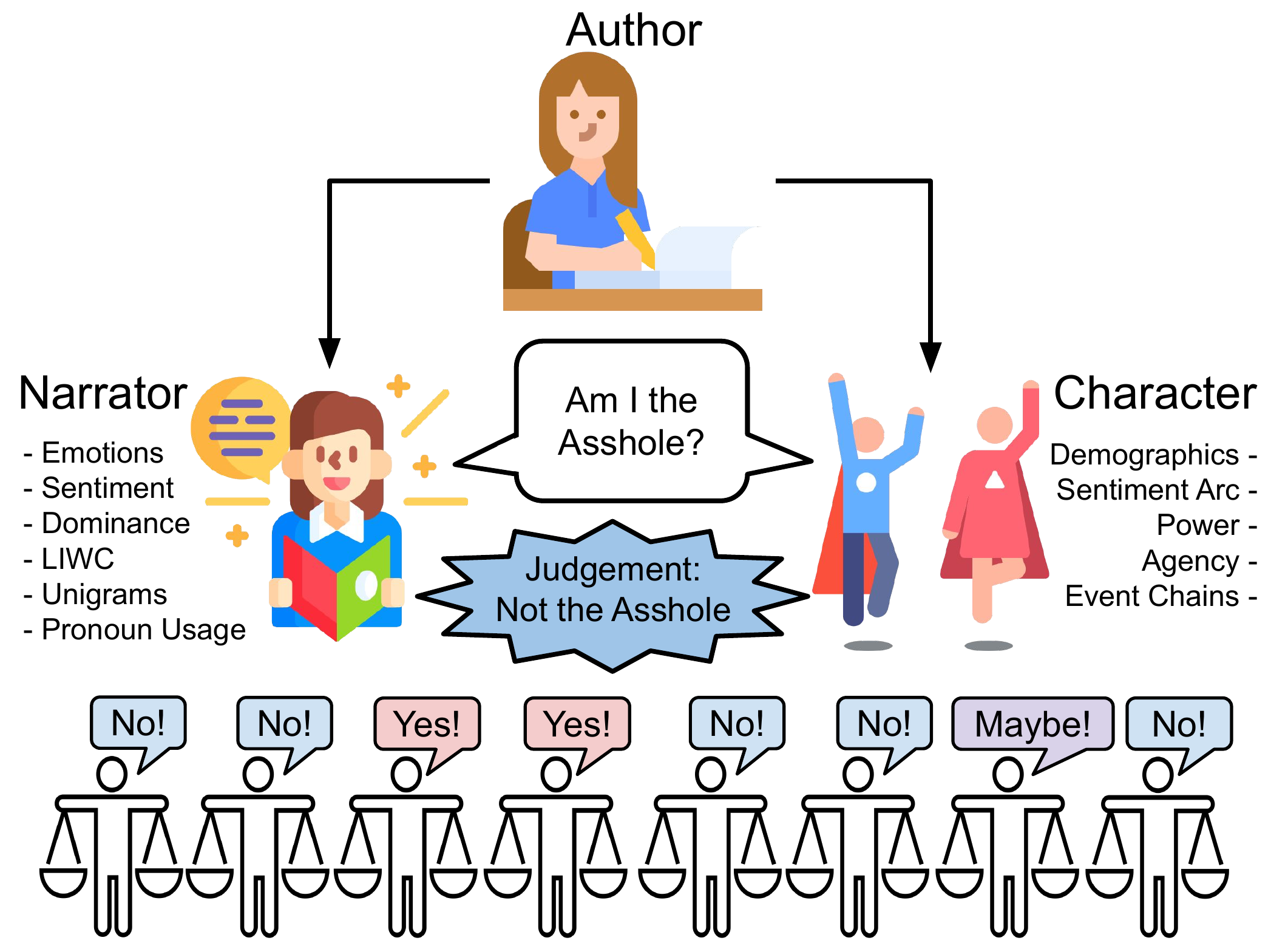}
  \caption{Crowd judgments of both the author as the narrator and a story character.} 
  \label{fig:flow}
\endminipage
\end{figure}

In this study, we look at anonymous, autobiographic tales (i.e., first person narratives, or personal narratives) from the \AITA~community.
In \AITA, people post their stories to help them determine (or convince themselves) that they have the moral high ground. Other people---who we will refer to as the Commenters---will vote on whether they believe the Original Poster (OP)\footnote{``OP'' and ``author'' will be used interchangeably throughout the paper. This is not to be confused with ``narrator,'' which we are considering to be the style of the written work.} is in the right (labeled ``Not the Asshole'', or \NTA~for short) or in the wrong (``You're the Asshole'', \YTA) and occasionally give their reasoning why.
Then the final label is selected through majority voting.

When looking at stories with a deeply-intertwined character/narrator dichotomy, such as in the \AITA~stories, it is worth noting how the moral judgment of the OP can change when the story is read from a different perspective.
%\todo{We should change this because we don't actually answer it here: Does the moral judgment of the story change when you read it from the perspective of the author as the narrator of the story vs a character within the story?}
Sagae et al., 2013~\cite{Sagae2013} consider the separation of the ``diegetic'' (the aspects of the story itself) and ``extradiegetic'' (the influence of the narrator) levels vital to understanding the amount of subjectivity in a story.
While narratives are known to be more persuasive than non-narratives, the mechanisms behind this are less well-understood~\cite{bullock2021narratives}. 
Previous research has suggested that \emph{identification}---the adoption of the perspective of the character---aids in narrative persuasion~\cite{cohen2001defining} among other mechanisms.

In this study, we take a computational linguistics approach to disentangle both the character and narrator in personal narratives in order to understand their downstream effects on moral judgments.
Here we define the \textit{author-as-character} features as those that relate to the content of the story that refers to the author (e.g, sentences where the author is the subject or direct object), whereas the \textit{author-as-narrator} features relate to the overall style of the story.
As such, we look at various linguistic and narrative features in order to ask {\bf RQ-C: What makes an asshole character?} compared to {\bf RQ-N: What makes an asshole narrator?}

Our contributions are as follows:
\begin{enumerate}
\item the introduction of a set of features for separating the author as character and the author as narrator;
\item a framework for analyzing autobiographical (or personal) narratives; and
\item the analysis of these features on both parts of the RQ.
\end{enumerate}
\noindent This work also acts as one answer to Piper et al.'s (2021 \cite{piper-etal-2021-narrative}) call for more story understanding work from a narrative perspective using information processing techniques.

In the rest of the paper, we will go into more detail on the specifics of the \AITA~dataset and discuss related work that uses this (or similar) data and other work investigating identities within stories. We then describe our two sets of features for the Author-as-Character \& Author-as-Narrator dichotomy. Next, we perform a logistic regression and show and discuss the results for answering RQ-C \& RQ-N, respectively. We end with a discussion about possible next directions.

\section{The \AITA~Data Set}
\AITA---an English-language subreddit on Reddit---is a forum where users can anonymously share their personal experiences (i.e., first person narratives) of a particular event or series of events they have been blamed for or believe they should be blamed for. In these accounts, authors are encouraged to explain any details of the events that they deem necessary in order for other Reddit users to pass judgment: is the original writer an asshole (\YTA---You're the Asshole) or not the asshole (\NTA) in the story? The other Reddit readers (the Commenters) vote on it and leave comments explaining their reasoning. Occasionally, readers also ask the original poster questions to clarify certain points, where the original poster can respond. According to the forum rules, Commenters are asked to start their comment with one of the following (1) You're The Asshole (\YTA), (2) Not The Asshole (\NTA), (3) Everybody Sucks Here (\verb|ESH|), (4) No Assholes Here (\verb|NAH|), and (5) Not Enough Info (\verb|INFO|). In the end, after 18 hours, each post is officially labeled by the tag corresponding to its top comments.
%Using this approach, users who publish the post would get a sense of their moral standings. 

Our data consists of scraped content from the \AITA~subreddit pulled from the Pushshift Reddit Data set~\cite{baumgartner2020pushshift}. The raw data is a collection of the initial posts of over 959,996 Reddit threads from June 2013 to June 2021. 
Since Reddit posts may stay online after deletion or the user has deleted their account, The Pushshift Reddit Data set contains deleted entries.
Our data has been cleaned and filtered to remove deleted submissions and submissions with no text in the body of the post.
Submissions with no text in the body of the post are typically posts with a title only or a title plus some non-textual body (such as an image).

The posts contain metadata such as creation time and subreddit names, as well as comments, but do not include the original poster’s user handle or the post's actual content. 
We remove posts from bots and moderators. Bots were identified through multiple methods: (1) bots identified in the subreddits \emph{r/botwatch}, \emph{r/spambotwatch}, and \emph{r/markov\_chain\_bots}, (2) a manual inspection of frequent posters, and (3) a manual inspection of user handles which contain the substring ``bot''. Moderators were also identified by manually inspecting high frequency posters or accounts with ``mod'' or ``moderator'' in their posts or user handle. 
%The final label is ``determined by the subscribers who have both rendered judgement and voted on which judgement is best.'' 

The final judgment label was not available in the Pushshift Reddit Data set, and thus needed to be scraped. We were able to scrape labels for 216,318 submissions via Selenium\footnote{https://github.com/baijum/selenium-python}.
In this study, we focus on two primary labels (\YTA~and \NTA) and dropped all remaining submissions. 

Additionally, we only consider submissions with at least 500 words and 20 comments. These minimums, respectively, have been used for accurately measuring person-level constructs (see the feature set below which includes emotions and sentiment; Eichstaedt et al., 2021~\cite{eichstaedt2021closed}) and as a threshold for \AITA~submissions with high engagement~\cite{zhou2021assessing}. 
Our final dataset consists of 38,060 submissions: 29,111 \NTA~(76.5\%) and 8,949 \YTA~(23.5\%). 

% TODO add this url after 
% https://github.com/sjgiorgi/moral_judgements_in_aita
% and change language to "has been"
All data (including the features described below) will be anonymized and publicly released.\footnote{\url{https://github.com/<redacted>/<redacted>}}. In particular, unigram frequencies will be manually inspected and any possibly sensitive data (e.g., names) will be removed. To respect the privacy of Reddit users (who have the ability to delete their posts), the full text associated with each Reddit post will not be released. A unique identifier associated with each post will be released which will allow researchers to merge our data set with data from the Reddit API and other publicly available data sets.

\section{Related Work} 

Personal narratives differ structurally from fictional stories as they are more likely to be nonlinear but tend to contain more salient events \cite{Sap2022}.
Researchers have used various methods to understand these types of narratives, including
text processing tools, such as word embeddings \cite{Bartal2022}, topic modeling \cite{Lukin2016,nguyen2022mapping}, sentiment analysis \cite{Lukin2016}, or low-level features like part-of-speech tagging and tokenization \cite{Swanson2014}.
Others rely on annotations for classifying complex aspects of personal narratives, such as identifying the intention of the narrator \cite{Lukin2016,Fu2019} or where there is subjectivity \cite{Sagae2013,Tammewar2020}.
We have found others who use pre-existing narrative models from the psychology community \cite{Saldias2020}.

Although earlier work in computational analysis of personal narratives focused on blog posts \cite{Swanson2008,Sagae2013,Swanson2014}, recent work has largely focused on social media sites such as Twitter and Reddit.

%work related to use of reddit
Reddit is a large public, potentially anonymous social media site, which is organized into hundreds of thousands of forums (or ``subreddits") related to specific topics.
Reddit is a rich data source for a number of NLP and Computational Social Science tasks: asking a favor~\cite{althoff2014ask}, gender differences in discussions of parenting~\cite{sepahpour2022mothers}, political discussions~\cite{guimaraes2019analyzing,de2021no}, maternal health~\cite{gao2021topic}, narrative power in birth stories written by mothers~\cite{antoniak2019narrative}, and self-improvement~\cite{dong2021room}. 
Reddit has also extensively been used to study mental health~\cite{de2014mental}, including depression~\cite{tadesse2019detection}, non-suicidal self-injury~\cite{giorgi2022addictionlanguage}, and suicide~\cite{matero-etal-2019-suicide,zirikly2019clpsych}.

% AITA in particular
Due to its unique nature, a number of recent studies have used the \AITA~subreddit. Botzer et al. (2021~\cite{botzer2021analysis}) investigated how users provide moral judgments of others, finding that users prefer posts that have a positive moral valence. Similar work using \AITA~attempts to automatically classify ``reasonability'' of people's actions based on a retelling of events from a story using a number of social and linguistic features (e.g., up/down votes and sentiment; Haworth et al., 2021~\cite{haworth2021classifying}).
Efstathiadis et al., 2021~\cite{efstathiadis2021explainable} built BERT-based classifiers for submissions and comments and attempts to predict both the final label and the comment labels. 
\AITA~has also been included in studies on advice communities (which includes the \emph{r/relationships} subreddit; Cannon et al., 2022~\cite{Cannon2022}) and ethical judgments ~\cite{lourie2021scruples}.
Finally, a number of studies attempt to understand judgments of moral dilemmas, rather than focus on a classification task.
For example, Nguyen et al., 2022~\cite{nguyen2022mapping} used topic modeling with expert and crowd-sourced annotations to understand moral dilemmas, showing that pairs of topics (e.g., family and money) are informative.
Similarly, Zhou et al., 2021~\cite{zhou2021assessing} showed that the \NTA~label is associated with more use of 1st person passive voice.

% normative behavior in storytelling in general
It is worth noting that other work \cite{nahian2020learning, Forbes2020} has framed moral/ethical judgments in stories as normative or non-normative behavior, acknowledging that morality is often person or culture-dependent, which we can see in the variance of judgments in the \AITA~data set as well.

\section{Feature Extraction}

We will explore features that represent the author as a character in the story and the narrator of the story. 
To reiterate from before: \textit{Author-as-character} features relate to the author as a character within the content of the story.
\textit{Author-as-narrator} features relate to how the author narrates the story.
We will use both {\em theoretically-driven} and {\em open-vocabulary} features. The theoretically-driven features are interpretable features which we hypothesize will be related to the final submission label. On the other hand, open-vocabulary features do not correspond to any prior hypotheses and make use of large feature spaces (e.g., unigrams and LIWC categories).

\subsection{Text Preprocessing}
Since all lexical and dictionary features (i.e., NRC, LIWC, Concreteness and Familiarity, and unigrams) are calculated via bag-of-words approaches, we apply the same text preprocessing steps. Submissions are first normalized: white space is collapsed, all characters are set to lowercase, and non-UTF8 characters are removed.
We then extract unigrams from each submission using a tokenizer designed to capture the idiosyncrasies of social media text (e.g., emoticons, misspellings; Schwartz et al., 2017~\cite{schwartz2017dlatk}).
Dictionary scores are then computed as the weighted sum of the unigram frequencies per submission (where LIWC weights are set to 1).
All other features (power/agency, chain of events, emotional story arc) are processed within spaCy.

\subsection{Author as Character}
The following feature categories describe who the author is as a person, their relationship to other characters in the story, and the story progression.

\paragraph{Demographics.} 
Many of the stories in the \AITA~submission corpus contain demographics (i.e., age and gender) of the author. For example, a typical submission will contain a line like ``my sister (66f), my two cousins (24F) and (18M), and me (51F)'' which denotes the age and gender of the author and various characters in the story. We extract author demographics through a series of regular expressions. Gender is numerically encoded such that male is -1, gender neutral is 0, and female is 1\footnote{We realize that this is a very limited sense of gender and that reducing gender to a binary representation is problematic. That said, our data is limited in its gender representation.}. We set age and gender equal to zero when narrators do not disclose their own demographics. To control for setting non-disclosed demographics to zero we include two binary covariates (one for age and another for gender) which are set to 1 for all authors who do not disclose either age or gender. 
The demographic information gives us 4 features: {\bf OP Age}, {\bf OP Gender}, {\bf Other Character Age}, {\bf Other Character Gender}. Both {\bf Other Character Age} and {\bf Other Character Gender} are averaged over all non-author characters in the narrative.

\paragraph{Power and Agency.} 
This is the amount of power and agency the OP has as a character.
Here we use the Power and Agency frames developed by Sap et al., 2017~\cite{sap-etal-2017-connotation}, where power is defined as control over the world while agency is control over oneself. The frames consist of a list of labeled verbs (1,737 for power and 2,146 for agency) where the label denotes the direction of the power between the subject and the direct object.  For example, if ``X dreads Y'' then Y (the \emph{theme} or direct object) has power, and if ``X excludes Y'' then X (the \emph{agent} or subject) has power. To measure the power of author in each story, we use the spaCy dependency parser\footnote{https://spacy.io/api/dependencyparser} to extract subject-verb-object tuples. We use a standard list of 1st person pronouns (including common misspellings) to identify the author: i, i'm, mine, myself, me. The author's power score is positively incremented if (1) they are the subject and verb's power label is \emph{agent} or (2) if they are the object and the verb's power label is \emph{theme}. We then normalize the power score by the number of times the author was included in a subject-verb-object tuple. When operationalized in this way, negative power means that the non-author entities exert power over the author, while positive power means that the author exerts power over other entities. Examples of different power and agency combinations are given in the Appendix Table \ref{tab:power agency examples}.
This gives us two features: {\bf OP Power} over other characters and {\bf OP Agency}.

\paragraph{Emotional Story Arc.} 
This is the emotional arc of the author as the main character, defined as the sentiment flow or progression across the story (or, more specifically, sentiment across sequences of sentences). 
Here we follow the methods of Antoniak et al. (2019 \cite{antoniak2019narrative}) and consider sentiment across the narrative.
%The authors measured power differentials between groups of characters in the narrative (e.g., the mother, doctor, midwife, and baby), finding that nurses are more powerful than other characters in the stories and babies are the least powerful.
%
We used the VADER sentiment analysis tool~\cite{hutto2014vader} to compute sentence-level sentiment across all submissions as implemented in the NLTK Python package~\cite{bird2009natural}. We ignore sentences with 5 words or less (due to their noisy sentiment estimates) and we only look at sentences that include the author (i.e., exclude sentences that do not reference the author in the subject-verb-object tuple). We use a normalized sentiment score that ranges from -1 to 1, with negative numbers representing negative sentiment and positive numbers representing positive sentiment. Next, for each submission, we average the sentence-level sentiment across 10 sequential and equal-sized chunks. We calculate the slope of the arc to see how the slope relates to the \NTA/\YTA~labels. Finally, we visualize the arcs by averaging the sentiment for each of the 10 chunks for both labels \NTA~and \YTA~and plotting the resulting averages.
%We call this feature the {\bf Emotional Story Arc}.

\paragraph{Chain of Events.} 
This is the narrative event chain of the story for all characters, focused on the verbs of the story's sentences.
We generate \NTA/\YTA~narrative event chains similar to the methods outlined in Tambwekar et al. (2019 \cite{Tambwekar2019}). We first extract the sequence of events (stemmed verbs from each sentence). Then, for stories in each label type (\NTA~or \YTA), we calculate two components: 1) the depth of the verb in the story (how many sentences in), and 2) the frequency of the verb across all stories. We then normalize the verb depth by dividing it by its frequency to get the average depth of each verb across all stories. Using this value, we cluster the verbs using the Jenks Natural Breaks optimization technique \cite{jenks1971error} to group together verbs found in similar positions across the stories, and this is our event chain. The process is repeated for both \NTA~and \YTA~stories. 
For each submission, we extract all the verbs in the order they appear and compare this to both \NTA/\YTA~chains via Damerau-Levenshtein distance, which gives us an approximation of how similar this post is to either of our event chains. %\cite{levenshtein1966binary}.
Finally, we pick the label with the higher matching score.
We considered chains of length 3, 5, and 10, picking chains of 3 clusters for achieving the highest accuracy in label prediction.
%We call this feature the {\bf Chain of Events}.
Having this feature highly correlate with \NTA~or \YTA~ would mean that posts within the label show a similar sequence of events.

\subsection{Author as Narrator}
The following features are centered around analyzing the author's tone and word choice.

\paragraph{Pronoun Usage.}
This is the amount of 1st-person and 3rd-person pronouns in the story.
Pronoun Usage will be measured using the Linguistic Inquiry and Word Count (LIWC) dictionary~\cite{pennebaker2001linguistic}. Here we will measure both 1st and 3rd person pronouns and encode each submission as a ratio of the two. This will give us an estimate of the narrator's focus on the self vs. others in the story.
This gives us 5 features: {\bf 1st Person Singular}, {\bf 1st Person Plural}, {\bf 3rd Person Singular}, {\bf 3rd Person Plural}, and the {\bf 1st/3rd Person Ratio}, regardless of plurality.

\paragraph{Sentiment-NRC.} This is the quantity of positive and negative sentiment words found in the story, providing the narrator's tone. We measure it via the {\em Positive Sentiment} and {\em Negative Sentiment} categories in the NRC Word-Emotion Association Lexicon, a crowd-sourced, word-level lexicon (Emolex; Mohammad and Turney, 2013~\cite{mohammad2013crowdsourcing}).
This gives us the features {\bf Positive Sent} and {\bf Negative Sent}.

\paragraph{Emotions-NRC.} Using the NRC Hashtag Emotion Lexicon~\cite{mohammad2015using} we estimate Plutchik's eight basic emotions: anger, anticipation, disgust, fear, joy, sadness, surprise, and trust~\cite{plutchik1980general}. 
The NRC Hashtag emotion lexicon, which is a set of weighted words for each emotion category, was automatically derived over tweets with emotion hashtags (e.g., \emph{\#anger} and \emph{\#joy}). This gives us the features {\bf Anger}, {\bf Anticipation}, {\bf Disgust}, {\bf Fear}, {\bf Joy}, {\bf Sadness}, {\bf Surprise}, and {\bf Trust}.

\paragraph{Dominant Tone.} From the NRC Valance, Arousal, Dominance (VAD) lexicon~\cite{mohammad2018obtaining}, we will use the dominance dimension which consists of 20,000 weighted (between 0 and 1) English words. High-dominance words include ``powerful'' \& ``success'' while low-dominance words include ``frail'' and ``empty''. This lexicon has previously been used to study the dehumanization (i.e., negative evaluation of a target group) of LGBTQ people in news articles~\cite{mendelsohn2020framework}.

\paragraph{Concreteness and Familiarity.} We use the MRC Psycholinguistic
Database which includes a lexicon of 85,941 weighted words for estimating \textbf{Concreteness} and \textbf{Familiarity}~\cite{paetzold-specia-2016-inferring}. Concreteness is a measure of how much a word refers to a tangible entity, while Familiarity refers to how often a word is seen or heard. %This lexicon has previously been used to study customer satisfaction~\cite{packard2021concrete}. 

\paragraph{LIWC.} Linguistic Inquiry and Word Count (LIWC) dictionary, which consists of 73 manually curated categories (e.g., both function and content categories such as positive emotions, sadness, and pronouns; Pennebaker et al., 2015~\cite{pennebaker2015development}). LIWC is the most widely used dictionary in social and psychological sciences with over 8,800 citations as of April 2020~\cite{eichstaedt2021closed}. %Thus, it is our hope that this dictionary can aid in interpreting what makes the narrator an asshole or not. 
This is an example of an {\em open-vocabulary feature} since we are considering the entire feature space.

\paragraph{Unigrams.} Using the tokenizer described above, we extract unigrams for each submission. This results in a total of 84,781 unigrams. We removed any unigram which was not used in at least 380 (1\%) of the submissions. This produced a final set of 2,726 unigrams. This number is smaller than the total number of observations (38,060 submissions), which will help prevent model over-fitting. As with LIWC, this is an example of an {\em open-vocabulary feature} since it uses the entire feature space.

\begin{figure*}[ht!]
\begin{subfigure}[c]{.49\textwidth}
  \centering
  \includegraphics[width=0.8\linewidth]{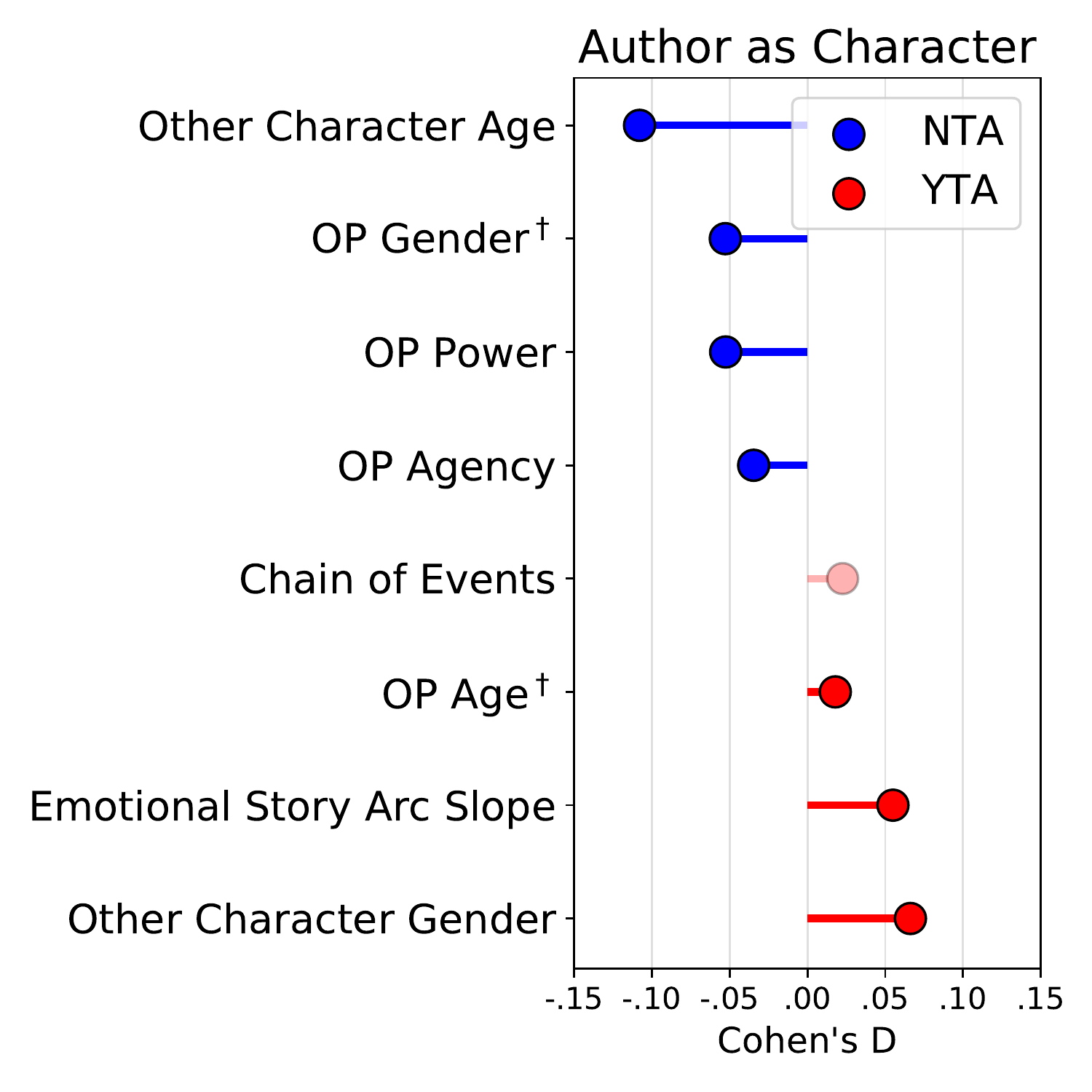}
  \caption{Cohen's D values showing the correlation of features for \YTA~\& \NTA~classes. Lighter shaded points are not significant at a with Benjamini-Hochberg corrected significance $\alpha<0.05$). The higher the absolute effect size, the more that feature is associated with the \YTA/\NTA~class. $\dagger$ includes a binary covariate equal to 1 for undisclosed age/gender.}
  \label{fig:character}
\end{subfigure}%
\hfill
\begin{subfigure}[c]{.49\textwidth}%[!b]
\centering
\includegraphics[width=\linewidth]{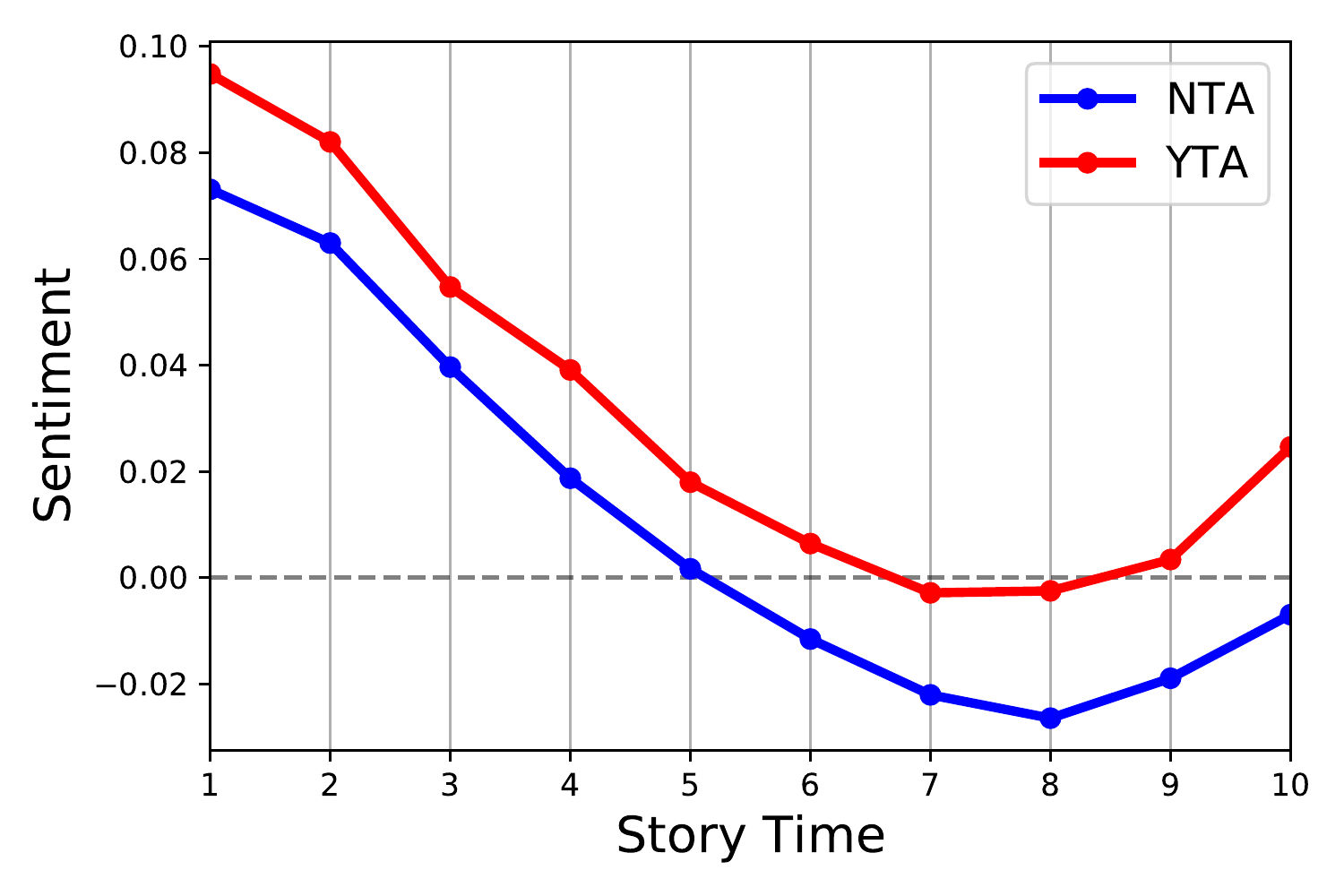}
\caption{Emotional Story Arc. Average VADER sentiment across 10 equally-sized sentence-level chunks. Positive values are positive sentiment, negative values are negative sentiment.} 
\label{fig:emotional arc}
\end{subfigure}
\caption{Author as Character results.}
\label{fig:character-both}
\end{figure*}

\section{The Makings of an Asshole}
%{\bf RQ1-Character: What makes an asshole character?}\\
%{\bf RQ1-Narrator: What makes an asshole narrator?}

\subsection{Methods}
In order to see which features in both sets (author as narrator and character) are indicative of a post being labeled \YTA, we run a correlational analysis.
These features will be used to quantify the content of the stories and how they related to the final vote of \YTA~or \NTA. For each feature we correlate, via logistic regression, the standardized feature value (mean-centered and normalized by their respective standard deviation) with the binary crowd-sourced label \YTA/\NTA, which is operationalized as 1 and 0, respectively. Due to a large number of comparisons (e.g., up to 2,726 unigrams), we apply a Benjamini-Hochberg False Discovery Rate correction~\cite{benjamini1995controlling}. 
Since logistic regression coefficients are not comparable across models in the same way OLS regression coefficients are, we report Cohen's D, which measures the difference between two group means (the \YTA/\NTA~classes). This is measured as the absolute difference in group means divided by the pooled standard deviation, with value close to 1 having a larger effect size. We note that Cohen's D is traditionally a positive number (due to the absolute difference). In order to aid interpretation, we set Cohen's D values for features correlated with the \NTA~class to be negative.

Due to shared methodologies across the feature space (e.g., the NRC lexica or the subject-verb-object tuples used to calculate power/agency and the emotional story arcs), we examine how the various features overlap in their relationship to the binary \YTA/\NTA~label. For each pair of features $f_1$ and $f_2$, we perform a logistic regression where the dependent variable is the binary \NTA/\YTA~label. Each regression contains three independent variables: two additive terms and a multiplicative term (the product of the two additive terms, i.e., the interaction term):
\begin{equation}
    y = \frac{exp\Big(\beta_0 + \beta_1 f_1 + \beta_2 f_2 + \beta_3\big(f_1\times f_2\big)\Big)}{1+exp\Big(\beta_0 + \beta_1 f_1 + \beta_2 f_2 + \beta_3\big(f_1\times f_2\big)\Big)} 
\end{equation}
When applicable, we also include the binary covariates for undisclosed author age and gender (discussed in the Feature Extraction section). All independent variables are standardized (mean-centered and divided by the standard deviation). Again, we apply a Benjamini-Hochberg False Discovery Rate correction since there are a large number of comparisons (18 total features). 

A significant interaction term $\beta_3$ indicates that the dependent variable and one of the additive independent variables depend on (or moderate) the other. Due to the symmetry of the equation, the moderator is typically chosen for theoretical reasons. Since we have no a priori hypotheses, we simply examine whether or not $\beta_3$ is significant and make no further claims on the relationships between $f_1$ and $f_2$. Due to the large feature space, in this analysis we only consider features that are significantly correlated with the \NTA/\YTA~label.

\subsection{RQ-C - Results \& Discussion}

\paragraph{Demographics.}
For the Character features in Figure \ref{fig:character}, we see the strongest correlation of younger and female authors associated with the \NTA~class (or similarly, older and male authors associated with the \YTA~class).
Stories with other characters who are younger (on average) 
and female (on average) are associated with \YTA; or similarly, stories with other characters who are older, male are associated with \NTA.  Note that for author age and gender, Cohen's D cannot take into account the binary variables (in the logistic regression) used to indicate undisclosed demographics. When the author's age or gender is undisclosed, we set the value to zero, which will drive Cohen's D toward 0, thus artificially deflating the effect size. 

\paragraph{Power and Agency.}
Both high author power (control over the world) and high author agency (control over themselves) are associated with the \NTA~class. Conversely, this could mean that \YTA~posters are having events happen to them.

\paragraph{Emotional Story Arc.}
The arc shapes in Figure \ref{fig:emotional arc} are often called a ``riches to rags` or ``tragedy'' story arc~\cite{reagan2016emotional}. While both arcs are similar, the \YTA~arc is consistently higher than the \NTA~arc (i.e., more positive on average), slightly crossing the 0 threshold (i.e., neutral or slightly negative sentiment), and has a larger upswing at the tail-end of the story. Meanwhile, the \NTA~arc, while following the same general shape, has a more neutral and negative sentiment overall. At each point, we compute a two-sided t-test (difference in means across the two classes) and see that the \NTA and \YTA are significantly different, with an average $t=6.07$ across the 10 story points. Thus, while the shapes are similar, there is a significant difference between the overall sentiment level at each point. The arc significantly interacts with other character's age and gender as well as the emotion \textbf{Disgust} (Table \ref{tab:interactions}).

In Figure \ref{fig:character} we also see more a positive Emotional Story Arc \emph{slope} associated with the \YTA~class. 
Given the overall ``riches to rags'' story shapes in Figure \ref{fig:emotional arc} slopes downwards (i.e., a decline in sentiment), a more positive slope for the \YTA~class would imply a less dramatic decline or even a constant or upward sentiment arc. 
Research on cinematic tragedies has shown that those who experience greater lows during tragedies also experience increased highs at the end of the tragedy~\cite{de1995role}. 
One possible explanation is that these increased highs or transitions from low to high may increase engagement with the \NTA~posts, though further research is needed to examine this relationship with moral judgments. 

The Chain of Events features are not significantly correlated with the final label after applying the FDR correction. The stories are most likely too unique to have the events signify \NTA~or \YTA.

% \YTA~is more positive -- vain?
% \NTA~eventually gets into negative
% The LIWC correlations are shown in Table \ref{tab:liwc correlations}, which includes the top five most positively and negatively correlated categories. We see mentions of family, home, time, and self-focus are more associated with the \NTA~label, while the insight, positive emotions, swear, and tentative categories are more associated with the \YTA~label. 

\subsection{RQ-N - Results \& Discussion}
\label{sec:RQ1-narrator}
\begin{figure*}[ht!]
\begin{subfigure}[c]{.49\textwidth}
  \centering
  \includegraphics[width=0.8\linewidth]{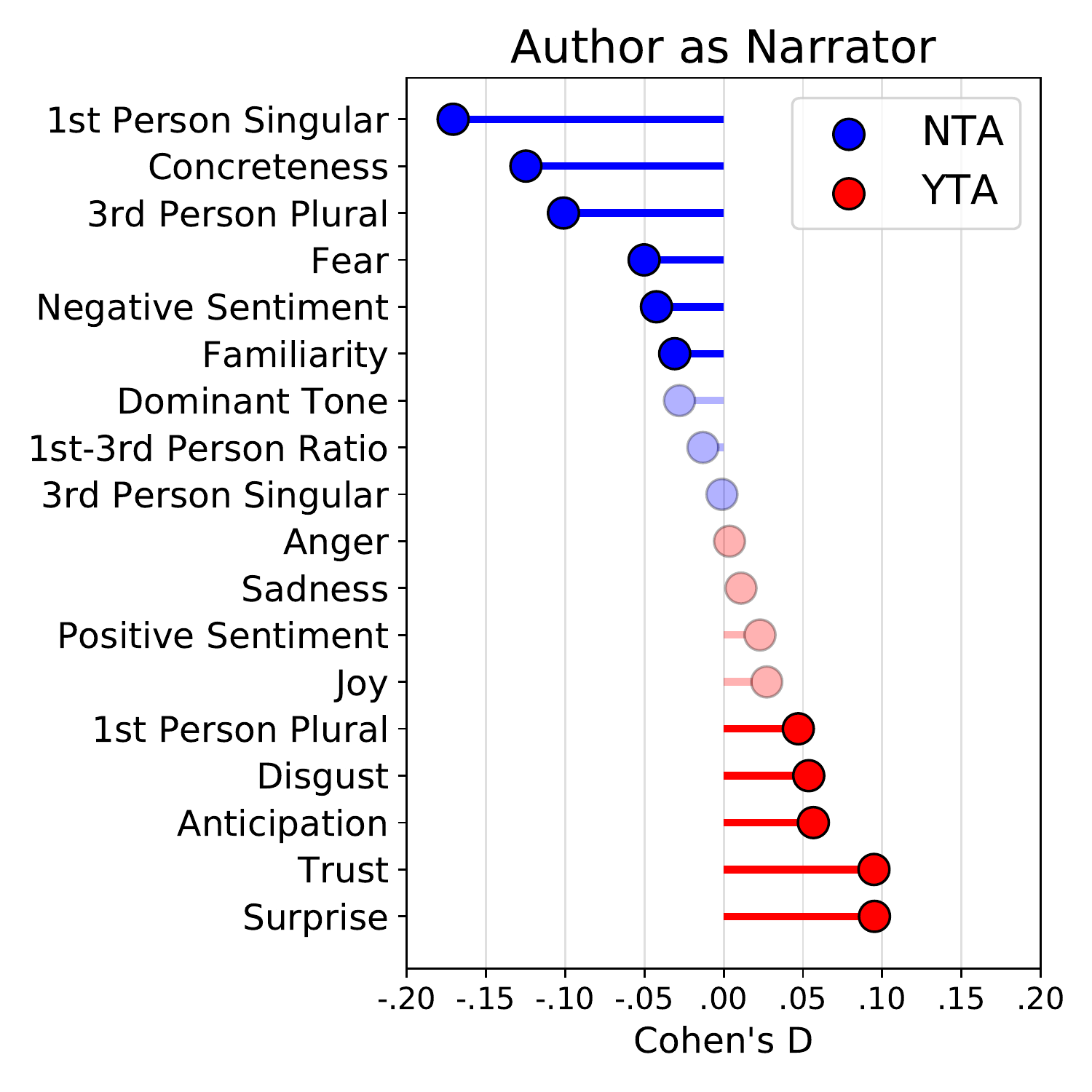}
  \caption{Lighter shaded points are not significant at a with Benjamini-Hochberg corrected significance $\alpha<0.05$).}
  \label{fig:narrator}
\end{subfigure}%
\hfill
\begin{subfigure}[c]{.49\textwidth}%[!b]
\centering
\resizebox{\columnwidth}{!}{
\begin{tabular}{p{.2cm}lcc}
\toprule
 & Category & Most Frequent Words & D\\  \hline
\multirow{5}*{\rotatebox{90}{\textbf{\textcolor{blue}{\NTA}}}} &{\em Family}  & mom, family, dad, mother, son       & -.24$^{***}$ \\
&{\em I}  &  i, my, me, i'm, i've & -.17$^{***}$ \\
&{\em Home}    & family, home, house, room, bed & -.17$^{***}$ \\
&{\em Pers. pronouns}  &  i, my, she, me, her & -.16$^{***}$ \\
&{\em Male}  &  he, him, his, dad, he's & -.14$^{***}$ \\ \hline
\multirow{5}*{\rotatebox{90}{\textbf{\textcolor{red}{\YTA}}}} &{\em Insight} & know, feel, think, thought, decided &  .22$^{***}$ \\
&{\em Cognitive Proc.} & but, not, if, because, all &  .22$^{***}$ \\
&{\em Differentiation} & but, not, if, or, really & .19$^{***}$ \\
&{\em Tentative} & if, or, some, any, anything &  .18$^{***}$ \\
&{\em Impers. pronouns} & that, it, this, what, other &  .17$^{***}$ \\
 \bottomrule
\end{tabular}}
\caption{LIWC Correlation results. Benjamini-Hochberg corrected significance levels: $^{***}$ p < 0.001, $^{**}$ p < 0.01, $^{*}$ p < 0.05. We also report the top 5 most frequent words in each category.}
\label{tab:liwc correlations}
\end{subfigure}
\caption{Author as Narrator results. Cohen's D values showing the correlation of features for \YTA~\& \NTA~classes for LIWC categories (\ref{tab:liwc correlations}) and other theoretically-driven Narrator features (\ref{fig:narrator}).}
\label{fig:narrator-both}
\end{figure*}

\begin{figure*}[ht!]
\begin{subfigure}{.5\textwidth}
  \centering
  \includegraphics[width=.8\linewidth]{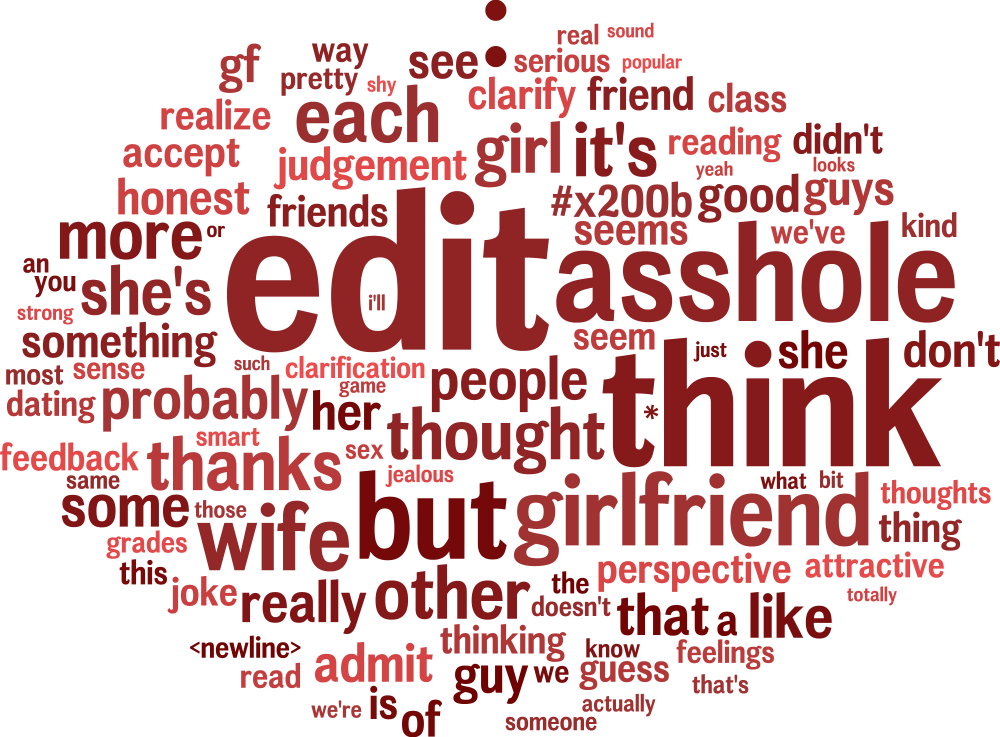}
  \caption{\YTA-Correlated Unigrams}
  \label{fig:sub-pos-unigram}
\end{subfigure}%
\begin{subfigure}{.5\textwidth}
  \centering
  \includegraphics[width=.8\linewidth]{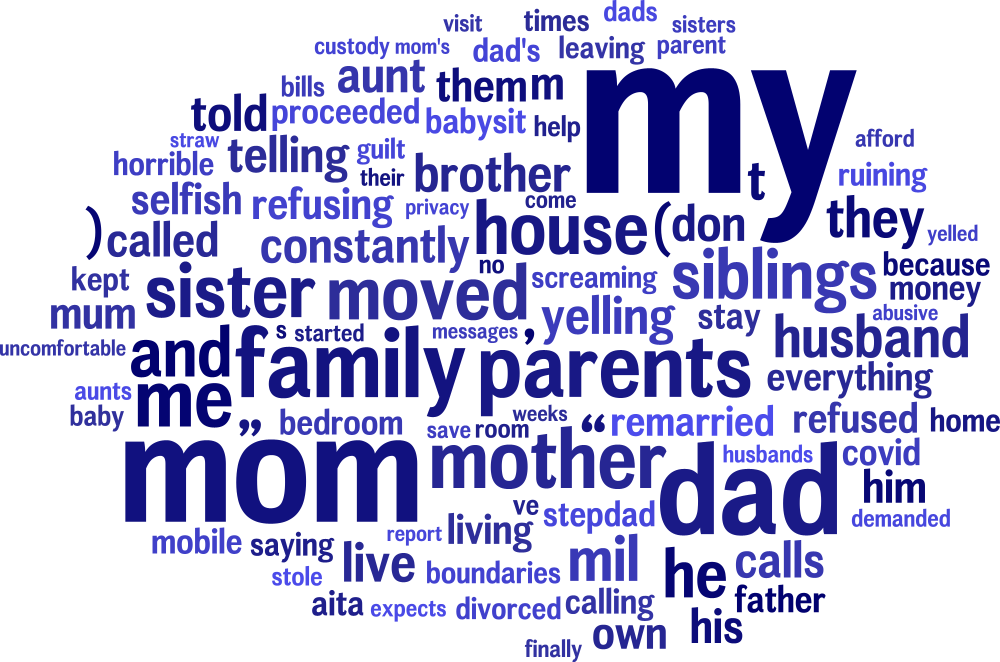}
  \caption{\NTA-Correlated Unigrams}
  \label{fig:sub-neg-unigram}
\end{subfigure}
\caption{Author as Narrator, open-vocabulary features: Unigrams most correlated with the (a) \YTA~label (left and (b) \NTA~(right). The size of the word indicates the strength of the correlation (large sizes correspond to larger Cohen's D); color indicates the relative frequency of usage (rare words are lighter shades, very frequent words darker). All correlations are significant at a Benjamini-Hochberg significance level of $p < 0.05$. Cohen's D ranged from (a) 0.06 to 0.24 and (b) -0.06 to -0.27 (see Appendix Table \ref{table:unigram correlations} for full Cohen's D value by unigrams). }
\label{fig:word-clouds}
\end{figure*}

Figures \ref{fig:narrator-both} \& \ref{fig:word-clouds} show the results of our analyses of features corresponding to the Author as Narrator, and whether or not they correlate with \YTA~or \NTA~labels.
%We admit that we found many of the results for the Author-as-Narrator features surprising and counter-intuitive at first.
%Here we see more self-focus (more first person singular pronouns) and higher dominance predicting the \NTA~class. Narrator features that most predict the \YTA~class include swear words, negative sentiment, and negative emotions.
%More ``I''/``me'' usage was highly correlated with \NTA, while third-person singular pronouns were correlated with \YTA. We had assumed we would see the opposite results---as a sign of OP's self-centeredness when labeled \YTA. Instead, we believe these results indicate accountability of OP in \NTA~stories and less shifting the blame onto other people.
%Similarly, more use of 1st person plural pronouns was correlated with being labeled \YTA, more 3rd person plural pronouns was correlated with \YTA, and a higher 1st/3rd person ratio correlated with \NTA.

\paragraph{Pronoun Usage.}
More ``I'' usage was highly correlated with \NTA. We had assumed we would see the opposite results---as a sign of OP's self-centeredness when labeled \YTA. Instead, we believe these results might show Commenters the accountability of OP in \NTA~stories.
\NTA~posts are also more likely to use 3rd person plural pronouns, but \YTA~posts use 1st person plural pronouns more. 

Tausczik and Pennebaker, 2010~\cite{Tausczik2010} have found that 1st person plural pronouns are a sign of having high status (i.e., ``the royal we'') or being detached, while 1st person singular pronouns are a sign of honesty and depression and 3rd person singular pronouns show social interest. We believe that, by using 1st person plural pronouns, \YTA~posters could be seen as detaching themselves from what they know to be a bad situation or thinking highly of themselves, while \NTA~posters are seen as more honest about their account and caring about the others in the story because of their pronoun usage.
We see similar patterns with the LIWC data in Figure \ref{tab:liwc correlations}: mentions of distinct entities (i.e., {\em Family} terms, personal pronouns ({\em I}, {\em Personal pronouns}, and {\em Male}) are more correlated with \NTA~posts. 

In terms of interactions (Table \ref{tab:interactions}), first person singular pronouns and first person plural pronouns significantly interact. Additionally, first person singular pronouns interact with \textbf{Trust} (negative coefficient) and first person plural pronouns interact with \textbf{Familiarity} (positive coefficient). Thus, referring to singular others who the narrator trusts increases the probability of the \NTA~class, while familiarity and first person plural pronouns increase the probability of the \YTA~class.

\paragraph{Sentiment-NRC \& Emotions-NRC.}
\textbf{Fear} and \textbf{Negative Sentiment} are more correlated with \NTA~posts. 
Zhou et al., 2021~\cite{Zhou2021} found that stories with emotional themes of ``distress'' or ``sadness'' garnered more empathy from readers. 
This could explain the association of \textbf{Fear} and \textbf{Negative Sentiment}, with Commenters empathizing with the Narrator and labeling the post as \NTA. That said, \textbf{Sadness} was not significantly correlated with either label.
Other emotions, such as \textbf{Disgust}, \textbf{Anticipation}, \textbf{Trust}, and \textbf{Surprise} are all more correlated with \YTA~posts. 
This could imply \YTA~stories relying more on emotional persuasion to get Commenters on their side and less on factual descriptions of the events of the story.

\paragraph{Concreteness and Familiarity.} 
Both \textbf{Concreteness} and \textbf{Familiarity} are correlated with the \NTA~label. 
Past research has shown that using abstract language to describe others is perceived to have biased motivations when compared to more concrete language~\cite{douglas2006you}.
This may suggest that \YTA~posts are perceived as having biased or hidden motives.

It has also been shown that online engagement depends on the complexity of the language: people spend more time and pay more attention to simple language; however, they are also more likely to give money if complex language is used, such as in grant proposals or crowdfunding~\cite{markowitz2021predictive}. 
This may explain the association between \textbf{Familiarity} and \NTA, as Commenters may spend more time reading and pay closer attention to \NTA~posts.
Further, \textbf{Familiarity} interacts with \textbf{Anticipation} (negative logistic regression coefficient) and, thus, increases the probability of the \NTA~class (Table \ref{tab:interactions}). 
\textbf{Concreteness} and \textbf{Familiarity} significantly interact increasing the probability of the \YTA~class.

\paragraph{LIWC.}
The results we found with the LIWC categories (Figure \ref{tab:liwc correlations}) were more predictable. In addition to personal pronouns, \NTA~posts also mention specific {\em Home} locations, while \YTA~posts use more {\em Insight} words (i.e. opinion words; see Figure \ref{tab:liwc correlations} for examples), {\em Tentative} words, and {\em Impersonal Pronouns}. This backs up our theory that \NTA-rated stories have a tendency for having more concrete reports, while \YTA~OPs will be vaguer.
Toma, 2014~\cite{Toma2014} saw that people perceive more information to be more trustworthy when looking at the Facebook profiles of strangers. They believed this to be because of Uncertainty Reduction Theory (URT)---the theory that people need more information about each other in order to decrease the amount of uncertainty \cite{Knobloch2015}.

\YTA~posts will also give more opinions of events in the story with {\em Cognitive Processes} and {\em Differentiation} words. This could be because they either feel as if the events of the story could not be understood without some extra context or that they did not hold up on their own.
Regardless of their intentions, using more opinion words can create narrative distance. Andringa1996, (1996~\cite{Andringa1996}) has seen that more opinionated narrators can make readers feel less ``emotional involvement'' in a story. Perhaps Commenters with less interest in stories are more likely to vote \YTA.

\paragraph{Unigrams.}
The most-associated unigrams for each label are shown in the word clouds in Figure \ref{fig:word-clouds} (see Appendix \ref{table:unigram correlations} for exact effect sizes). In Figure \ref{fig:sub-pos-unigram} (\YTA), we see the word ``edit'' largest effect size (Cohen's D). This means that the OP had gone back to change/add to their original story. This result goes along with \YTA~posters feeling the need to provide extra context. Anecdotally, across Reddit, OPs usually tend to edit their post when they are being criticized by Commenters and feel the need to elaborate or defend themselves.
\YTA~stories are more likely to mention ``asshole'' directly.

There is also more mention of romantic and sexual relationships: ``dating'', ``attractive'', and ``sex'', especially with a focus on words referring to women: ``girlfriend'', ``gf'', ``wife'', ``she's'', ``girl''. Although we do not know the exact context that these words are used, we suspect them to denote sexist undertones in the post, especially if the OP is more likely to be older and male if tagged \YTA~(see: OP Gender \& OP Age results in the Author-as-Character analysis). Very few mentions of men are in the \YTA~word cloud, with the exception of ``guy''/``guys''.

We also see %words which may refer to poor attempts at humor (``joke'') or bluntness (``honest'', ``serious'') and 
``thinking'' words (``think'', ``thought'', ``perspective'', ``feedback'', ``judgement'', ``realize'', ``admit''), which can be seen as attempts to understand what they (might have) done wrong or could be more evidence that opinionated posts create too much narrative distance (see: LIWC discussion).

In Figure \ref{fig:sub-neg-unigram} (\NTA), we see mentions of family (``mom'', ``family'', ``dad'', ``siblings''), self-focus (``my'' and ``me''), conflict (``refused'', ``horrible'', ``yelling'', ``demanded''), as well as financial and health problems (``bills'', ``money'', ``covid'') and stressful life-changing events (``moved'', ``divorced'', ``custody''). We also see open and close parenthesis, which could indicate parentheticals containing extra information and details given by the narrator, backing up the URT theory.
\NTA~posts also contain more quotation marks, which either indicate direct quotes or disagreement with the word or phrase someone else has said.

Dominant Tone was not statistically significant.

\subsection{Discussion Summary}

Now we will summarize the results in order to answer our research questions.

{\bf RQ-C: What makes an asshole character?}
Typically, if the OP is older and male with younger and female other characters in the story, the author is considered an asshole, probably due to an uneven power dynamic. These OP characters, which lack both agency and power, are written as having events {\em happen to them}. One possible explanation is that the author is framing their character as the victim. Meanwhile, the events that include them are generally positive, implying that good things are happening to their characters in the story.

{\bf RQ-N: What makes an asshole narrator?}
An ``asshole narrator'' is more detached from the story but tells more emotional and opinionated stories with fewer concrete facts all-the-while framing stories as a matter of perspective. The distance that the OP creates through this style makes Commenters less invested in the narrative. These posts might also be seen as less truthful. Furthermore, ``asshole narrators'' focus on topics such as relationships with women instead of family or stressful situations.

\section{Limitations and Future Work}

We admit that a lot of our conclusions about what the results may mean are conjecture until further analysis is done.
However, we believe that our methods of separating out the narrator's style from the character's actions are helpful for understanding autobiographical narratives, as the narrator is not just an impartial observer. This linguistic breakdown (or similar) might be used in situations such as identifying con artists online, quickly determining who reliable narrators are in humanitarian crisis live-tweeting, or in narrative criminology.

There is a limitation to how much we can tease apart the author as narrator from the author as character given that they are both presented in the same medium: text. We attempted to treat the problem as style vs. content, but we admit some of the lines are unclear. For example, we treat OP's gender as a character feature, but does the OP's gender also affect how they write their stories?

Furthermore, we only ran our analysis on a single data set---the \AITA~subreddit. It is possible that there are other data sets, such as the similarly-themed {\em r/relationship\_advice} subreddit, that could also work for both analyzing first-person narratives and comparing it to moral judgments from external parties. Other data sets can also be created using first-person narratives with crowd-sourced moral judgements~\cite{lourie2021scruples}.

All models used throughout this paper (LIWC, NRC lexica, Vader) were originally built on monolingual English datasets, and we have applied them in the same setting. Therefore, we do not expect any findings to generalize to non-English or minority populations. 

Although we used several different features for analyzing posts from the perspective of the author as the narrator or character, we admit that there could be features that we did not think of. Our features are not an exhaustive list of all possible ways of characterizing these posts. 

One major source of data that we have not analyzed, for instance, is the comments left on the posts. Future work could compare the text from the comments to the events or wording of the story in order to determine what readers are actually paying attention to when making their moral judgments. 
%Future work could focus on comments, where one could attempt to identify which authors' roles the Commenters pay attention to when defending their moral judgements. 
Style matching has been shown to predict author credibility~\cite{aune1993effects}, and thus one might expect Commenters to match the narrator in emotions or function words.
This, however, we deemed out of the scope of this paper and we leave for future work.

Future work could also expand on the analytic framework for measuring positioning in narratives as developed by Kayi-Aydar  (2021~\cite{kayi2021framework}) which attempts to formalize how narrative identities are constructed, projected, or negotiated. Finally, one could examine narrative framing and related causal structures. For example, one could take these posts, modify the power direction of the verbs, and reevaluate the story in a crowd-sourced setting (like Amazon Mechanical Turk) to see if the final \NTA/\YTA~label changes.

\section{Broader Perspective and Ethical Concerns}

While the methods in this paper are evaluated on a single data set, \AITA, we believe the general concept of separating the author-as-narrator from the author-as-character is potentially useful across several domains. From a computational perspective, those working in narrative understanding or character extraction could build on the methods here~\cite{bamman-etal-2013-learning,jahan-finlayson-2019-character}. From a social science perspective, political scientists and those working in media communications could be interested in disambiguating the author in the context of narrative persuasion~\cite{braddock2016meta} or how narratives shape public opinion (a situation comparable to asking ``who is the asshole?'')~\cite{card-etal-2016-analyzing}. 

When working with public social media data there are always a number of ethical concerns. While \AITA~subreddit is a public forum where users are requesting moral judgments from their online peers, it is important to note that the Redditors have not consented to any research studies. Indeed, this problem is not particular to \AITA~and is part of a larger issue of using publicly available social media data in research. While focused on mental health applications, Chancellor et al., 2019~\cite{chancellor2019human} consider who is the ``human'' in machine learning research that uses social media data and discuss a number of implications around informed consent. As such, to preserve anonymity, all results are reported in aggregate, and we do not report direct quotes.

Also of note is the fact that we use age and gender to classify moral judgments, including a very narrow (binary) definition of gender. We do not intend to imply that any given age or gender is or should be considered an ``asshole.''

\section{Conclusion}
In this paper, we have distinguished two facets of the author in first person narratives: the Author as Character and the Author as Narrator, and quantify what makes each an asshole. To do this, we automatically extracted a large number of interpretable linguistic features designed to measure story characters and events as well as narrative tone and style. We performed a correlational analysis to give insight into which character and narrator features are related to overall moral judgments. Among other things, we found that asshole characters are older and male who have events passively happen to them, while asshole narrators frame the story as a matter of perspective. We believe by considering the narrator as separate from their character in the story, we are able to get a deeper insight into not just how these narratives are stylized but also how the author writes themselves into the story. Future work should be done to understand how readers understand these two aspects of autobiographical stories.

% for camera ready version
% The authors declare no competing interests.

\bibliographystyle{unsrt}  
\bibliography{anthology,references}

\section{Appendix}

\subsection{Power and Agency examples}

Table \ref{tab:power agency examples} includes example sentences which for all combinations of high / low power and high / low agency. Note that all posts have been paraphrased and anonymized for privacy.

\subsection{Unigram Correlations}
In Table \ref{table:unigram correlations} we list the exact Cohen's D values for the top 10 most correlated unigrams for each label.

\subsection{Interactions}
In Table \ref{tab:interactions} we report the standardized logistic regression coefficients associated with the multiplicative interaction logistic regression term. 

\begin{table}[!h]
    \centering
    \begin{tabular}{p{.9\columnwidth} }%{cl}
    \toprule
        \textbf{Positive Agency \& Theme Power} \\ \hline
        I \hlc[pink]{asked} her many times over the last few weeks to make up her mind, but she can't.  \\ \hline
        \textbf{Positive Agency \& Agent Power} \\ \hline
        Because of this and the fact that people were staying to help her i \hlc[pink]{decided} to leave. \\ \hline
         \textbf{Negative Agency \& Theme Power} \\ \hline
         My friend \hlc[pink]{needed} the money in order to stay out of debt. \\ \hline
         \textbf{Negative Agency \& Agent Power} \\ \hline
         Let me talk the way i \hlc[pink]{want}! \\
        \bottomrule
    \end{tabular}
    \caption{Examples of sentences with all combinations of agent (subject) / theme (object) power and positive / negative agency. All posts have been paraphrased and anonymized.}
\label{tab:power agency examples}
\end{table}

\begin{table}[h]
    \centering
\begin{tabular}{cccc} \toprule
\multicolumn{2}{c}{\YTA} & \multicolumn{2}{c}{\NTA} \\ \cmidrule(lr){1-2}\cmidrule(lr){3-4}
Unigram & Cohen's D & Unigram & Cohen's D \\ \hline
edit & 0.24 & my & -0.27 \\
think & 0.20 & mom & -0.19 \\
: & 0.16 & dad & -0.16 \\
asshole & 0.16 & family & -0.13 \\
but & 0.15 & parents & -0.13 \\
girlfriend & 0.13 & mother & -0.12 \\
wife & 0.12 & me & -0.12 \\
each & 0.11 & sister & -0.11 \\
it's & 0.11 & moved & -0.11 \\
thanks & 0.11 & house & -0.11 \\ \bottomrule
\end{tabular}
\caption{Cohen's D values for the top unigrams associated with each label.}
\label{table:unigram correlations}
\end{table}

\begin{table}[hb]\centering
\resizebox{\columnwidth}{!}{
\begin{tabular}{lcccccccccc} \toprule
 & \multicolumn{3}{c}{Character} & \multicolumn{7}{c}{Narrator} \\ \cmidrule(lr){2-4}\cmidrule(lr){5-11}
 & (1) & (2) & (3) & (4) & (5) & (6) & (7) & (8) & (9) & (10) \\ \hline
(1) Emo. Story Arc  & - & .06 & .06 & ns & ns & ns & ns & .04 & ns & ns \\
(2) Other Char. Age & .06 & - & ns & ns & ns & ns & ns & ns & ns & ns \\
(3) Other Char. Gen. & .06 & ns & - & ns & ns & ns & ns & ns & ns & ns \\
(4) 1st Pers. Plur. & ns & ns & ns & - & .09 & ns & ns & ns & .05 & ns \\
(5) 1st Pers. Sing. & ns & ns & ns & .09 & - & ns & ns & ns & ns & -.04 \\
(6) Anticipation & ns & ns & ns & ns & ns & - & ns & ns & -.03 & ns \\
(7) Concreteness & ns & ns & ns & ns & ns & ns & - & ns & .02 & ns \\
(8) Disgust & .04 & ns & ns & ns & ns & ns & ns & - & ns & ns \\
(9) Familiarity & ns & ns & ns & .05 & ns & -.03 & .02 & ns & - & ns \\
(10) Trust & ns & ns & ns & ns & -.04 & ns & ns & ns & ns & - \\ \bottomrule
\end{tabular}
}
    \caption{Pairwise interactions: Reported standardized logistic regression coefficient of the multiplicative interaction regression term. \emph{ns}: interaction term is not significant at $p < 0.05$ after a Benjamini-Hochberg False Discovery Rate correction. Due to a large number of comparisons, features are not included in the table if they had no significant interactions with any other feature.}
\label{tab:interactions}
\end{table}

\end{document}